\documentclass[10pt,twocolumn,letterpaper]{article}

\usepackage{iccv}
\usepackage{times}
\usepackage{epsfig}
\usepackage{graphicx}
\usepackage{amsmath}
\usepackage{amssymb}
\usepackage{caption}
\usepackage{subcaption}
\usepackage{enumitem}
\usepackage{booktabs}
\usepackage{adjustbox}
\usepackage{pifont}
\usepackage{multirow}

\usepackage[accsupp]{axessibility}
\usepackage[toc,page]{appendix}
\newcommand{\xmark}{\ding{55}}%
\newcommand{\RN}[1]{%
  \textup{\uppercase\expandafter{\romannumeral#1}}%
}
\usepackage{color, colortbl}
\definecolor{Gray}{gray}{0.9}	
\definecolor{LightCyan}{rgb}{0.88,1,1}

\usepackage[pagebackref=\checkmark,breaklinks=\checkmark,letterpaper=\checkmark,colorlinks,bookmarks=false]{hyperref}

\iccvfinalcopy 

\def\iccvPaperID{20} 

\ificcvfinal\fi
\begin{document}

\title{Masking Strategies for Background Bias Removal in Computer Vision Models}

\author{Ananthu Aniraj\thanks{Corresponding Author} $^{1,3,4,5}$  \quad Cassio F. Dantas$^{2,3,5}$ \quad Dino Ienco$^{2,3,5}$ \quad Diego Marcos$^{1,3,4,5}$
\\ \\
 $^{1}$Inria \quad $^{2}$Inrae \quad $^{3}$University of Montpellier \quad $^{4}$LIRMM \quad $^{5}$UMR-Tetis  \\
{\tt\small \{ananthu.aniraj, diego.marcos\}@inria.fr}
\quad
{\tt\small \{cassio.fraga-dantas, dino.ienco\}@inrae.fr}
}

\maketitle
\ificcvfinal\thispagestyle{empty}\fi

\begin{abstract}
Models for fine-grained image classification tasks, where the difference between some classes can be extremely subtle and the number of samples per class tends to be low, are particularly prone to picking up background-related biases and demand robust methods to handle potential examples with out-of-distribution (OOD) backgrounds. 
To gain deeper insights into this critical problem, our research investigates the impact of background-induced bias on fine-grained image classification, evaluating standard backbone models such as Convolutional Neural Network (CNN) and Vision Transformers (ViT).
We explore two masking strategies to mitigate background-induced bias: Early masking, which removes background information at the (input) image level, and late masking, which selectively masks high-level spatial features corresponding to the background. Extensive experiments assess the behavior of CNN and ViT models under different masking strategies, with a focus on their generalization to OOD backgrounds.
The obtained findings demonstrate that both proposed strategies enhance OOD performance compared to the baseline models, with early masking consistently exhibiting the best OOD performance. Notably, a ViT variant employing GAP-Pooled Patch token-based classification combined with early masking achieves the highest OOD robustness. 

Our code and models are available at: \href{https://github.com/ananthu-aniraj/masking_strategies_bias_removal}{GitHub}


\end{abstract}


\section{Introduction}

\begin{figure}[t]
\centering
    \includegraphics[width = 0.98\linewidth]{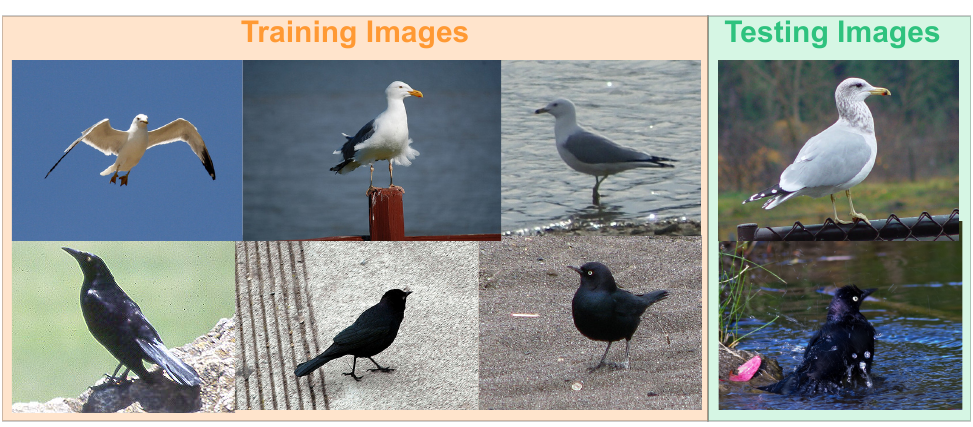}
    \caption{Background bias in fine-grained classification }
    \label{fig:prob_statement}
\end{figure}
The context in which an object is found is known to have a large influence in how it is perceived, both by humans~\cite{biederman1982scene} and machines~\cite{choi2012context}.
When training deep learning-based computer vision models on any object-centric task, it is extremely challenging to ensure that the model does indeed only pay attention to the object of interest, since there exist often strong correlations between object characteristics and the background. Such a bias can lead to degraded performances when deploying the models on new, Out-of-Distribution (OOD), background configurations~\cite{barbu2019objectnet}. This is particularly harmful in fine-grained classification tasks such as species identification (\autoref{fig:prob_statement}), due to the large number of closely related categories and the strong correlations between the object (the specimen of interest) and the background, which tends to relate to the habitat of the species~\cite{beery2018recognition,geirhos2020shortcut, xiao2020noise}. 

The standard way to address this type of bias is to make the model detect the foreground and focus on it for classification, both when using Convolutional Neural Networks~\cite{luo2021rectifying} and Transformers~\cite{hiller2022rethinking}. This reduces the influence of the background on the model's decision, thus reducing the background-induced bias of the model. 

Masking the background areas at some CNN high level representations seem like a simple and straight forward strategy to obtain this effect, as long as some type of foreground-background (FG-BG) mask can be obtained. After all, the inductive biases that characterize CNNs will induce an association between a high-level feature at a certain tensor location (i.e. before any global pooling) and the image patch at the corresponding image location. However, the Visual Transformers' architecture does not encode for such biases, although they could be learned via supervised or self-supervised learning~\cite{caron2021emerging,oquab2023dinov2}.

\begin{figure*}[t]
\centering
    \includegraphics[width = 0.86\textwidth]{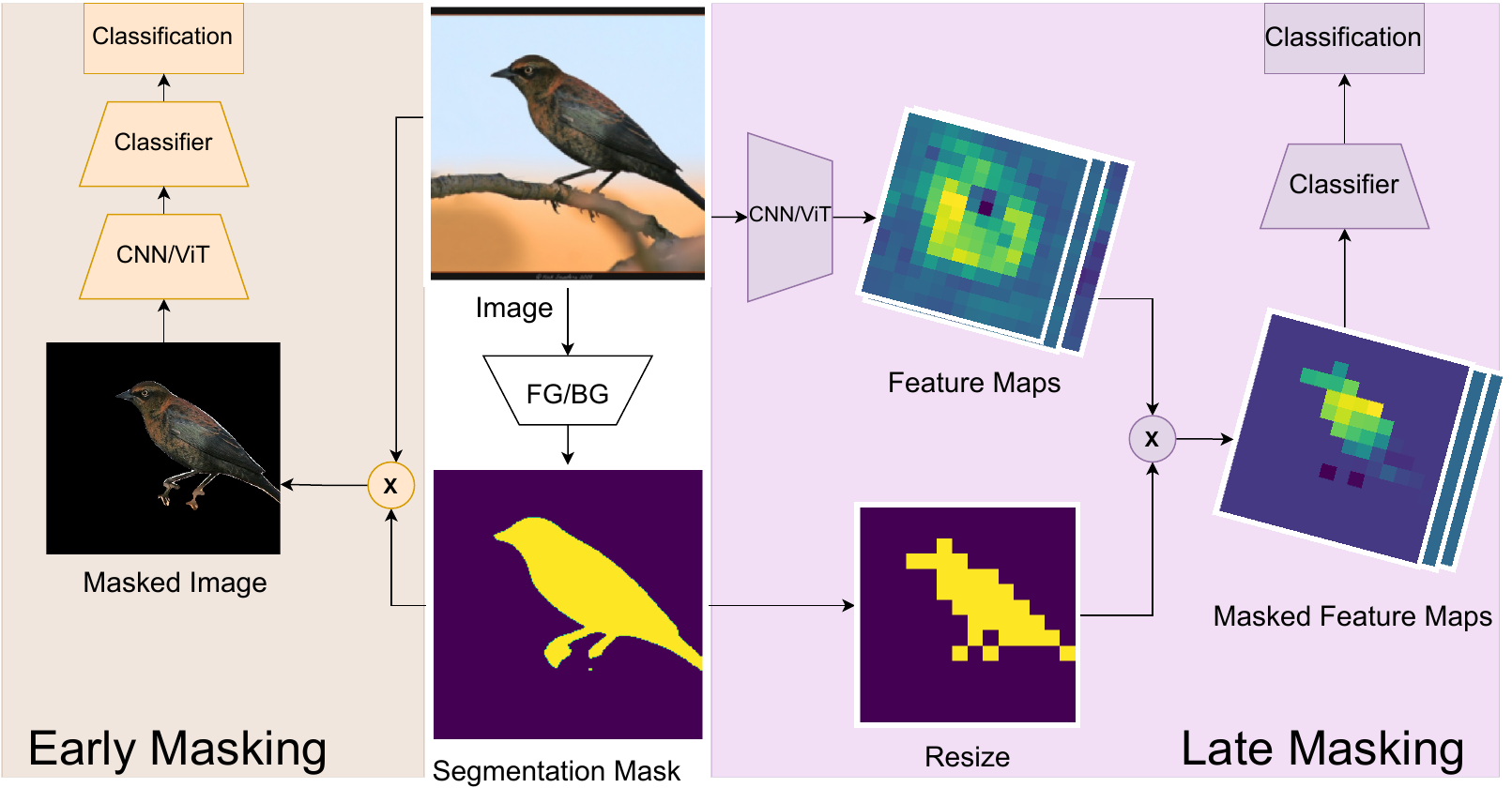}
    \caption{Early and late masking strategies}
    \label{fig:em-lm-approach}
\end{figure*}

We aim to analyse the impact of dataset bias induced by background regions in the image on the performance and generalization capabilities of image classification models on fine-grained classification tasks using standard CNN and Vision Transformer (ViT) \cite{Dosovitskiy2021AnScale} backbones. 
Specifically, we study the effect of background masking, both at the image level and at high level features, on both CNN and ViT, on a fine-grained classification task. 
Our main motivation is to understand the difference in behaviour, in terms of generalization to OOD backgrounds, of ViT with respect to CNN under different strategies of background masking.





\section{Related Work}

\paragraph{The influence of context} The ability of using context for object recognition is often considered an asset, since the context may help disambiguate between otherwise indistinguishable categories, and the explicit modeling of context played an important role in pre-deep learning computer vision systems~\cite{choi2011tree,choi2012context}. 
With the emergence of deep learning approaches, which are capable of automatically extracting features without leveraging user knowledge, the extraction of spurious correlations induced by contextual bias has started to be perceived as a potential issue~\cite{geirhos2020shortcut,singh2020don,xiao2020noise}. 
Such effect can happen to the extent that CNN models may perform reasonably well when given only the background, without the object of interest~\cite{zhu2017object}.

\paragraph{Mitigating context-induced bias} In early computer vision systems for object recognition, performing a pre-processing step of foreground segmentation was identified as an effective solution against background bias~\cite{rosenfeld2011extracting}.
When the context classes are known, another approach is to make sure that the learned attention maps for each class do not overlap~\cite{singh2020don}.
Alternatively, the manifold of learned class features can be explored in order to manually identify spurious ones that can then be suppressed~\cite{neuhaus2022spurious}.
With modern Transformer-based architectures, it has been shown that allowing the model to refocus its attention in a top-down manner (thus based on the results after a first pass) helps in a variety of computer vision tasks~\cite{shi2023top} by reducing the effect of background elements.

In this work, we study the effect on OOD background generalization of masking the image background regions, at either the image level or at the level of an intermediate representation within the model.




\section{Methodology}

We aim at studying the effect of dataset bias introduced by the image background regions and to what extent it can be corrected via background masking. We propose to investigate the following three training settings:

\begin{itemize}
    \item \textbf{Baseline:} \label{it:base-line} Fine-tune a standard image classification model from the literature on a fine-grained image classification dataset.
    \item \textbf{Early masking:} \label{it:early-mask} Apply background masking at the image level using foreground-background (FG-BG) masks. We then fine-tune image classification models to classify the masked out image.
    \item \textbf{Late masking:} \label{it:late-mask} Mask out high-level representations that spatially correspond to background regions at an intermediate stage of an image classification model using the FG-BG masks. Subsequently, we fine-tune the resulting model.
\end{itemize}


The FG-BG masks are obtained by applying a semantic segmentation network trained on ground-truth foreground-background masks. 

In order to investigate the generalization capabilities of the models under the different settings, we propose to evaluate them on a test set that has a different background distribution than the train set.

\subsection{Binary FG-BG Segmentation}
For binary segmentation, we trained a deep neural network using the segmentation labels provided by the CUB-200-2011 dataset \cite{WelinderEtal2010}. The objective was to classify each pixel in the image as either bird or background.

Given an image $\mathbf{x}\in\mathbb{R}^{3\times M\times N}$, the segmentation network was trained to produce a binary mask $\mathbf{m}\in\{0,1\}^{1\times M\times N}$, where $0$ represents the background, and $1$ represents the foreground.


\subsection{Early Masking}

For this strategy (\autoref{fig:em-lm-approach} left), we first mask out the background image regions using the binary segmentation network and then pass the masked image to a standard CNN/Transformer-based image classifier model. The CNN/ViT model is composed of two stages: a backbone $h_{\theta_1}(\cdot)$ (CNN or ViT) and a classification head $g_{\theta_2}(\cdot)$. More formally, we can define it as:

\begin{equation}
    \mathbf{y} = g_{\theta_2}(h_{\theta_1} (\mathbf{x} \odot \mathbf{m}) )
\end{equation}
where each channel of the image ($\mathbf{x}$) is multiplied element-wise by the FG-GB binary mask ($\mathbf{m}$) and $\mathbf{y}\in\mathbb{R}^C$ is the vector of class logits.

The idea here is to mask out background features at the image level, forcing the model to only look at the salient object in the image for the classification task, ideally learning representations independent of bias caused by the image background.


\subsection{Late Masking}

In this strategy (\autoref{fig:em-lm-approach} right), the high-level spatial features corresponding to the background are masked out, similar to the feature masking layer proposed in~\cite{Dai2015ConvolutionalSegmentation}. More formally, this strategy can be defined as:

\begin{equation}
\mathbf{y} = g_{\theta_2}(h_{\theta_1} (\mathbf{x})\odot \mathbf{m}')
\end{equation}

Here, $\mathbf{y}$ represents the final classification logits, $\mathbf{x}$ is the input image, and $\mathbf{m}'$ is the subsampled version of the binary FG-BG mask.

The backbone $h_{\theta_1}(\cdot)$ outputs a spatial tensor $\mathbf{z}$.
\begin{equation*}
\mathbf{z}=h_{\theta_1}(\mathbf{x})\in\mathbb{R}^{D\times M'\times N'}
\end{equation*}
where $M'=M/k$ and $N'=N/k$, with $k$ being the model's spatial downsampling ratio. In the case of ViT, this spatial tensor results from the spatial rearrangement of the patch tokens' features, each a vector of dimension $D$.

The classification head $g_{\theta_2}(\cdot)$ performs a global average pooling followed by at least one learnable linear layer.

The late masking strategy allows selectively masking out background-related features at a later stage in the model, enabling the model to learn representations based solely on foreground image features.

\section{Experiments}

\subsection{Dataset}
\label{subsec:datasets}

All our models are trained on the training set of CUB \cite{WelinderEtal2010}. This dataset contains images of 200 bird species, with 5,994 images in the training set and 5,794 images in the testing set.

The trained models are evaluated on two test sets: (i) the in-distribution real-world images of the CUB test set \cite{WelinderEtal2010} and (ii) an Out-Of-Distribution (OOD) set of images with an adversarial background, specifically the Waterbirds dataset from~\cite{sagawa2019distributionally}.

The Waterbirds dataset~\cite{sagawa2019distributionally} was constructed by replacing the background regions in the CUB images with ones coming from the Places dataset \cite{zhou2017places}. Therefore, this dataset contains the exact same bird species as CUB, with the only difference being the background.

\subsection{Implementation Details}
\label{subsec:implementation_details}

All models were implemented in PyTorch. For the sake of comparison, we used the ImageNet-pretrained ConvNeXt \cite{Liu2022A2020s} and DinoV2-pretrained ViT \cite{oquab2023dinov2} backbones as starting weights for all the experiments.

We utilized the Small (S), Base (B), and Large (L) variants of the ViT and ConvNeXt models for our experiments. These variants allowed us to explore the impact of model size on the performance of our masking strategies.

To perform binary segmentation, we fine-tuned a pre-trained Mask2Former \cite{cheng2021mask2former} model with a Swin-Tiny backbone \cite{liu2021Swin} using the FG-BG masks provided along with the CUB dataset. We use only the images from the CUB training dataset to train the segmentation model. The Mask2Former model was chosen for its effectiveness in image segmentation tasks. However, in theory, any semantic segmentation model can be used in its place. Detailed information regarding the training settings for the binary segmentation model, as well as the evaluation results, can be found in \autoref{FirstAppendix}.

\begin{table}[ht]
\centering
\begin{adjustbox}{width=0.95\linewidth}
\begin{tabular}{@{}lll@{}}
\toprule
Models                 & ConvNeXt-S/B/L &    ViT-S/B/L    \\ \midrule
Input Size             & (224, 224)     & (518, 518)   \\
Optimizer              & AdamW  \cite{Loshchilov2019DecoupledRegularization}         & AdamW  \cite{Loshchilov2019DecoupledRegularization}       \\
Base LR     & 4e-3           & 1e-3         \\
Weight Decay           & 5e-2           & 1e-2         \\ 
Optimizer momentum     & (0.9, 0,999)   & (0.9, 0,999) \\
Batch Size             & 128            & 128          \\
Training Epochs        & 90             & 90           \\
Learning Rate Schedule & Cosine Decay   & Cosine Decay \\
Warmup Epochs          & None           & None         \\
Warmup Schedule        & N/A            & N/A          \\
RandAugment \cite{9150790}            & (9, 0.5)       & (9, 0.5)     \\
Mixup \cite{Zhang2018MixUp:Minimization}                  & None           & None         \\
Cutmix \cite{Yun2019CutMix:Features}                 & None           & None         \\
Random Erasing \cite{Zhong2020RandomAugmentation}         & 0.25           & 0.25         \\
Label Smoothing \cite{muller2020does}        & 0.1            & 0.1          \\
Stochastic Depth \cite{10.1007/978-3-319-46493-0_39}       & 0.1/ 0.2/ 0.3    & 0.0          \\
Frozen Backbone         & \checkmark          & \checkmark        \\
\bottomrule
\end{tabular}
\end{adjustbox}
\caption{Training Settings for frozen backbone} 
\label{tab:train_settings_exp2_frozen}
\end{table}

\begin{table}[ht]
\centering
\begin{adjustbox}{width=0.95\linewidth}
\begin{tabular}{@{}lll@{}}
\toprule
Models                 & ConvNeXt-S/B/L &    ViT-S/B/L    \\ \midrule
Input Size             & (224, 224)     & (518, 518)   \\
Optimizer              & AdamW \cite{Loshchilov2019DecoupledRegularization}          & AdamW \cite{Loshchilov2019DecoupledRegularization}        \\
LR(Backbone)     & 4e-6           & 4e-6         \\
LR(Classifier)     & 4e-3           & 4e-3         \\
Weight Decay           & 5e-2           & 5e-2         \\ 
Optimizer momentum     & (0.9, 0,999)   & (0.9, 0,999) \\
Batch Size             & 64            & 64          \\
Training Epochs        & 300             & 300           \\
Learning Rate Schedule & Cosine Decay   & Cosine Decay \\
Warmup Epochs          & 20           & 20         \\
Warmup Schedule        & Linear            & Linear         \\
RandAugment \cite{9150790}            & (9, 0.5)       & (9, 0.5)     \\
Mixup \cite{Zhang2018MixUp:Minimization}                  & 0.8           & 0.8         \\
Cutmix \cite{Yun2019CutMix:Features}                 & 1.0           & 1.0         \\
Random Erasing \cite{Zhong2020RandomAugmentation}         & 0.25           & 0.25         \\
Label Smoothing \cite{muller2020does}        & 0.1           & 0.1          \\
Stochastic Depth \cite{10.1007/978-3-319-46493-0_39}       & 0.1/ 0.2/ 0.3    & 0.1          \\
Frozen Backbone & \xmark & \xmark \\
\bottomrule
\end{tabular}
\end{adjustbox}
\caption{Training Settings for fine-tuning} 
\label{tab:train_settings_exp2_finetune}
\end{table}
\subsection{Training Settings}
\label{subsec:train_settings}

For all our experiments, we used two different versions of training settings:

\begin{enumerate}[noitemsep,nolistsep, label=(\roman*)]
\item For models with a frozen backbone, we utilized the settings described in \autoref{tab:train_settings_exp2_frozen}. In this case, only the final classification layer was trained, and we employed a relatively short training schedule of 90 epochs.
\item For models with fine-tuning, we employed the settings described in \autoref{tab:train_settings_exp2_finetune}. In this case, all layers in the network were trained for a longer schedule of 300 epochs, with additional regularization techniques like MixUp \cite{Zhang2018MixUp:Minimization} and CutMix \cite{Yun2019CutMix:Features}.
\end{enumerate}

\begin{table*}[ht]
\centering
\begin{tabular}{cc|cc|cc}
\toprule
 &  & \multicolumn{2}{c}{CUB(\%)} & \multicolumn{2}{c}{Waterbird(\%)} \\ \midrule
 & Training on CUB &  Test on original & Test on masked &  Test on original & Test on masked\\ \midrule
\multirow{2}{*}{ConvNeXt-S} &Baseline & \textbf{86.56} & 78.60 & 55.82 & 67.12 \\
& Masked & 83.84 & 84.43  & 64.21  & \textbf{77.33} \\ \midrule
\multirow{2}{*}{ConvNeXt-B} & Baseline  & \textbf{88.00}  & 77.52 & 65.96 & 68.51 \\
& Masked & 84.36 & 85.69  & 67.21 & \textbf{80.10} \\ \midrule
\multirow{2}{*}{ConvNeXt-L} & Baseline  & \textbf{88.05} & 78.68 & 66.83 & 70.69 \\
& Masked & 87.22 & 87.38  & 73.67 & \textbf{82.81}  \\ \midrule
\multirow{2}{*}{ViT-S} & Baseline & \textbf{88.26}  & 82.27 & 71.15 & 77.59 \\
& Masked & 85.89 & 87.92  & 78.61 & \textbf{84.05}  \\ \midrule
\multirow{2}{*}{ViT-B} & Baseline & 89.20  &  87.69 & 76.65 & 83.00 \\
& Masked & 88.35 & \textbf{90.10}  & 82.15 & \textbf{86.93}  \\ \midrule
\multirow{2}{*}{ViT-L} & Baseline & 89.79 & 88.91 & 80.76 & 85.61 \\
& Masked & 88.35 & \textbf{91.06} & 84.67 & \textbf{88.30}  \\ \bottomrule
\end{tabular}
\caption{Results: Early Masking vs Baseline }
\label{tab:results_exp1}
\end{table*}

\subsection{The Effect of Early Masking and Model Size}
\label{subsec:exp1}

The main aims of this experiment are to investigate: 
\begin{enumerate}[noitemsep,nolistsep,label=(\roman*)]
    \item The impact of adopting our early masking strategy.
    \item The interplay between this masking strategy and varying model sizes.
\end{enumerate}

For this experiment, we train classifiers while keeping the ViT and ConvNeXt backbones of three different sizes frozen. The training is performed on the CUB dataset, with two scenarios: one involving input images masked using the predicted FG-BG masks and the other without any masking. To extract features for classification in the ViT model, we concatenate the class token with the global average-pooled (GAP) patch token features, following a similar implementation as described in \cite{oquab2023dinov2}.

To ensure a comprehensive evaluation, we also assess the model trained on masked images with the original images and vice versa. This provides insights into the effectiveness of the early masking strategy and its sensitivity to different model sizes.

\subsection{Baseline vs Early Masking vs Late Masking}
\label{subsec:exp2}

The aims of this experiment are to study:
\begin{enumerate}[noitemsep,nolistsep,label=(\roman*)]
    \item The effects of applying both of our proposed masking strategies.
    \item The performance with a frozen and a fine-tuned backbone
\end{enumerate}

For this experiment, we chose two models of approximately the same size: the ViT-Base and the ConvNeXt-Base, due to their balanced compromise between performance and computational cost.

\subsection{Feature masking at different model stages}
\label{subsec:exp_sl}
The main aim of this experiment is to study the effect of applying late feature masking at different stages in the network. In this scenario, early masking would be equivalent to applying masking at stage $L=0$.

Concretely, we experimented with performing the feature masking after the second-to-last stage of the model ($L-1$) and compared the OOD and in-distribution performance with the masking after the last stage ($L$) and the early masked model ($L=0$).

The training settings for this experiment are similar to the fine-tuning settings given in \autoref{tab:train_settings_exp2_finetune}. Here, the model training requires two learning rates: one before and one after the feature masking, to ensure convergence. 

We were only able to perform this experiment for the ConvNeXt models, as we encountered challenges in finding stable training hyperparameters for masking the ViT at the $(L-1)^{th}$ stage.

\subsection{Varying the ViT representation}
\label{exp:mod_vit_class}

In this experiment, we aim to:
\begin{enumerate}[noitemsep,nolistsep, label=(\roman*)]
\item Vary the input to the ViT classification head to explore different ablations of the information extracted by the ViT model.
\item Investigate the performance differences between these ablations when the backbone is frozen versus when the entire model is fine-tuned.
\end{enumerate}

More precisely, we conduct experiments and evaluate the performance using three different inputs for the linear classification layer: (i) using the CLS token, (ii) using the Global Average Pooled (GAP) Patch token, and (iii) concatenating them together (CLS+Patch).To manage computational constraints, we focus solely on the ViT-Base model for this study.



\section{Results and Discussion}
\label{sec:results_discussion}

\subsection{The Effect of Early Masking and Model Size}
\label{subsec:exp1_results}


The performance of models trained with a frozen backbone, both with and without the early masking strategy, is evaluated using all combinations of test settings. The results of these evaluations are reported in \autoref{tab:results_exp1}. 

From the table, it is evident that the models trained with the early masking technique consistently outperform their counterparts trained without the masking strategy on the OOD Waterbirds test set, regardless of the backbone and model size. This improvement holds true when the early-masked models are tested on both the original and masked images. For instance, with ConvNeXt-B, the OOD performance improves from 65.96\% to 80.10\% when the images are masked at both training and testing time, and for ViT-B, it increases from 76.65\% to 86.93\%.

These results suggest that models trained on the original images are influenced by biases from the image backgrounds, which leads to lower generalization on the Waterbird test set when the background is not masked.

We also observe a performance boost for the baseline models when tested on the early masked version of the Waterbirds Dataset. This suggests that these models might have been relying on background cues to make decisions, which the early masking strategy helps to mitigate.

Furthermore, the best overall performances on the original CUB dataset are obtained by ViT-L and ViT-B when early masking is performed both at train and test stage. In contrast, when using convolutional models, this setting performs less well than the original one. This suggests that: (i) ViT models are less sensitive than CNNs to potential artifacts induced by masking and (ii) the original CUB dataset is negatively affected by background-induced biases. 
\begin{table*}[ht]
\centering
\begin{tabular}{@{}cccccc@{}}
\toprule
                            &                    & \multicolumn{2}{c}{CUB(\%)} & \multicolumn{2}{c}{Waterbird(\%)} \\ 
Backbone & Training on CUB & \multicolumn{1}{l}{Frozen} & \multicolumn{1}{l}{Fine-tuned} & \multicolumn{1}{l}{Frozen} & \multicolumn{1}{l}{Fine-tuned} \\ \midrule
\multirow{3}{*}{ConvNeXt-B} & Baseline                & 88.00   & 89.62     & 65.96        & 76.20        \\
                            & Early-Masked         & 85.69     & \textbf{90.31}     & 80.10        & \textbf{87.01}       \\
                            & Late-Masked & 87.66     & 88.76     & 77.95        & 78.42        \\ \midrule
\multirow{3}{*}{ViT-B}      & Baseline                & 89.20     & 89.38     & 76.55        & 68.36        \\
                            & Early-Masked         & 90.10     & \textbf{91.37}     & 86.93        & \textbf{88.81}        \\
                            & Late-Masked & 88.61     & 90.73     & 76.55        & 74.76 \\ \bottomrule       
\end{tabular}
\caption{Results of \ref{subsec:exp2} - Frozen backbone vs Fine-tuning} 
\label{tab:results_exp2_finetune}
\end{table*}
\subsection{Baseline vs Early Masking vs Late Masking}
\label{subsec:exp2_results}

\subsubsection*{Effect of freezing the model backbone}

The results are given in \autoref{tab:results_exp2_finetune}. Upon analyzing the table, it is evident that the early masking strategy shows the highest generalization performance on the OOD Waterbird test set for both backbones.

For the ConvNeXt model with a frozen backbone, we observe a substantial performance improvement when employing both the early and late masking training strategies.  This suggests that the high-level features spatially corresponding to the foreground are highly localized \cite{Luo2016UnderstandingNetworks}, likely due to the inductive biases in the CNN architecture, resulting in high spatial correlation of features even at the later stages of the network.

In the case of the ViT model with a frozen backbone, the late-masked and baseline models perform similarly on both the in-distribution CUB dataset and the OOD Waterbird dataset. This indicates that the early masking strategy helps prevent the ViT model, with frozen backbone parameters, from being influenced by background-related biases.

\subsubsection*{Effect of fine-tuning}

Fine-tuning consistently improves in-distribution performance on the CUB test set for both ConvNeXt and ViT models (\autoref{tab:results_exp2_finetune}). However, we do not observe a similar behavior on OOD data.

The early-masked model, for both ViT and CNN, demonstrates the best generalization, consistent with the findings from previous experiments.

For ConvNeXt, fine-tuning the baseline and early masked models leads to improvements in both in-distribution performance and significant advances in OOD generalization. However, the late-masked model shows only marginal performance gains.

In the case of the ViT, we see a reduction in robustness for both the baseline and late-masking variant on the OOD Waterbirds test set, with the baseline experiencing a more evident performance drop-off. This indicates that, with longer training, the ViT model becomes more susceptible to overfitting when background correlations are present in the training data, unless the background is masked out at the image level.

\subsection{Feature masking at different model stages}
\label{subsec:exp_sl_results}

\begin{table}[ht]
\centering
\begin{tabular}{cccc}
\hline
Backbone & \begin{tabular}[c]{@{}c@{}}Feature \\ Masking\end{tabular} & CUB(\%) & Waterbird(\%) \\ \hline
\multirow{3}{*}{ConvNeXt-S} & \textit{L}   & 88.73 & 77.19 \\
                            & $L-1$ & 89.35 & 81.22 \\
                            & \textit{0}   & \textbf{90.73} & \textbf{87.95} \\ \hline
\multirow{3}{*}{ConvNeXt-B} & \textit{L}  & 88.76 & 78.42 \\
                            & $L-1$ & 89.04 & 80.04 \\
                            & \textit{0}   & \textbf{90.31} & \textbf{87.01} \\ \hline
\multirow{3}{*}{ConvNeXt-L} & \textit{L}   & 89.80 & 79.39 \\
                            & $L-1$ & 89.49 & 79.68 \\
                            & \textit{0}   & \textbf{90.99} &   \textbf{88.19}\\ \hline
\end{tabular}
\caption{Results of adopting feature masking at different stages. In this table, L: after last stage, $(L-1)$: after second last stage. 0: early masking/at image level}
\label{tab:results_sl_fm}
\end{table}

The results of adopting feature masking at different stages in the network are presented in \autoref{tab:results_sl_fm}. It shows that both the in-distribution and OOD performance systematically improve when background masking is applied at an earlier stage in the network. This observation suggests that earlier layers of the CNN are more spatially correlated with the image patches at their corresponding locations, possibly due to the comparatively smaller receptive fields. 

Furthermore, we can note that the improvement in robustness of the models tends to decrease as the CNN model size increases, unless the masking is done at the image level itself. For instance, the ConvNeXt-Large model performs approximately the same regardless of whether the masking was applied after the last stage or the second last stage.

Overall, these results suggest that feature masking at an earlier layer of the network has a positive impact on both in-distribution and OOD performances.
\begin{table*}[ht]
\centering
\begin{tabular}{@{}ccccccc@{}}
\toprule
\multirow{2}{*}{Backbone} &
  \multirow{2}{*}{Training on CUB} &
  \multirow{2}{*}{\begin{tabular}[c]{@{}c@{}}ViT \\ representation\end{tabular}} &
  \multicolumn{2}{c}{CUB(\%)} &
  \multicolumn{2}{c}{Waterbird(\%)} \\
 &
   &
   &
  \multicolumn{1}{l}{Frozen} &
  \multicolumn{1}{l}{Fine-tuned} &
  \multicolumn{1}{l}{Frozen} &
  \multicolumn{1}{l}{Fine-tuned} \\ \midrule
\multirow{9}{*}{ViT-B} &
  \multirow{3}{*}{Baseline} &
  CLS &
  90.04 &
  \textbf{90.47} &
  80.35 &
  \textbf{71.35} \\
 &                               & Patch     & 62.24 & 90.05          & 24.71 & 67.74          \\
 &                               & CLS+Patch & 89.20 & 89.38          & 76.65 & 68.36          \\ \cmidrule(l){2-7} 
 & \multirow{3}{*}{Early-Masked} & CLS       & 90.33 & \textbf{91.73}          & 86.81 & 88.60 \\
 &                               & Patch     & 72.17 & 91.51 & 66.37 & \textbf{89.22} \\
 &                               & CLS+Patch & 90.10 & 91.37          & 86.93 & 88.81          \\ \cmidrule(l){2-7} 
 & \multirow{3}{*}{Late-Masked}  & CLS       & 89.16 & 90.78          & 75.10 & 80.54          \\
 &                               & Patch     & 84.15 & \textbf{90.85} & 71.63 & \textbf{84.50}  \\
 &                               & CLS+Patch & 88.61 & 90.73          & 76.55 & 74.76          \\ \bottomrule
\end{tabular}
\caption{Results of varying the representation of the ViT.} 
\label{tab:results_vit_classifier}
\end{table*}
\subsection{Varying the ViT representation}
\label{subsec:res_vit_cls_mod}

The results of the experiment are reported in \autoref{tab:results_vit_classifier}. Fine-tuning the models generally improves in-distribution performance on the CUB dataset. Notably, the model using only Patch tokens exhibits a substantial accuracy increase from 62\% to 90\% for the baseline, 72\% to 92\% for the early masked, and 84\% to 91\% for the late-masked variants.

On the OOD Waterbird dataset, fine-tuning the baseline model using both CLS and Patch tokens leads to reduced performance. This pattern also repeats for the model using just the CLS token. However, we see a notable improvement for the model using just the Patch tokens, where the accuracy increases from 25\% to 68\%.

In contrast, the early and late-masked models benefit from fine-tuning, showing improved OOD generalization with either CLS or Patch token input. Particularly, the ViT model employing GAP-Pooled Patch token classification, considering both early and late masking strategies, clearly outperforms all the other variants.

The experiment highlights the importance of studying which representations are utilized for the classifier layer in ViT models to enhance performance in fine-grained image classification tasks, particularly in the presence of background-induced biases. The use of Patch tokens, especially in conjunction with early or late masking, emerges as a promising strategy. This approach allows the model to focus on foreground-containing tokens and enhances model generalization with respect to background-induced biases.  

\section{Limitations and Future Work}
\label{sec:limitations}
Our study provides valuable insights and lays the foundation for potential solutions; however, certain limitations point to avenues for future exploration:

\textbf{Dataset Scope and Pre-trained Models:} The research study was mainly restricted to the CUB dataset, and we utilized publicly available pre-trained models due to time and computational constraints. Expanding our assessment to encompass larger datasets featuring diverse species and backgrounds, while also maintaining better control over pre-training methodologies, will enhance the generalizability and robustness of our conclusions.

\textbf{Real-World Robustness:} Our reliance on FG-BG ground truth annotations during segmentation model training may not align with scenarios where such privileged information is unavailable. Deploying our proposed models on real-world datasets without access to FG-BG annotations could yield further insights into its robustness.

\textbf{Computational Overhead:} Our proposed masking strategies involving the segmentation model introduce computational overhead, which could be infeasible in scenarios with tight time constraints. To this end, future research could explore techniques that seamlessly integrate background suppression as a training objective, aiming for enhanced out-of-distribution generalization. Notably, recent work by Chou et al.~\cite{Chou2023Fine-grainedSuppression} also emphasizes similar directions in the context of fine-grained image classification. 

These limitations underscore the potential for future research endeavors aimed at refining and expanding the effectiveness of our proposed methods.
\section{Conclusion}
\label{sec:conclusion}
In this study, we conducted extensive experiments to investigate the impact of background-induced bias on the generalization capabilities of CNN and ViT models for fine-grained image classification on out-of-distribution (OOD) data and explored various background masking strategies to mitigate such biases and enhance model generalization.


 Overall, our results revealed that the proposed masking strategies improve OOD performance, with the early masking strategy exhibiting the best generalization capability on both CNN and ViT. Early masking is particularly advantageous when fine-tuning the model backbones, with late masking not benefiting from fine-tuning and even loosing some OOD generalization capacity in the case of ViT.  Notably, a ViT variant employing GAP-Pooled Patch token-based classification, combined with early masking, exhibits the highest OOD robustness of all tested settings.
 
While our study provides valuable insights and solutions, certain limitations remind us of unexplored frontiers. Expanding investigations to larger datasets, controlled pre-training methodologies, and considering real-world scenarios lacking FG-BG annotations are essential steps toward refining and extending the impact of our methods.

In conclusion, our study emphasizes the significance of considering background information in object-centric tasks and demonstrates effective solutions to mitigate biases and enhance model generalization. 





{\small
\bibliographystyle{ieee}
\bibliography{references,egbib}
}

\appendix
\section{Binary Segmentation}
\label{FirstAppendix}
\subsection{Training Details}
We fine-tuned a semantic segmentation model pretrained on the ADE20k \cite{zhou2017scene} dataset on the FG-BG masks provided by the CUB dataset \cite{WelinderEtal2010}.

The training settings are exactly the same as the original Mask2Former paper \cite{cheng2021mask2former}. We tested models with the ResNet50 and the Swin-Tiny \cite{liu2021Swin} Backbones and trained both models for a total of 16000 epochs with the AdamW \cite{Loshchilov2019DecoupledRegularization} optimizer 

\subsection{Evaluation Metrics}
We used the Mean Dice Score \cite{ed278621-dc3e-343f-ae66-540d8990b60d} to evaluate the segmentation quality of the FG-BG segmentation models. 

The Dice score is calculated using the equation given below.
\begin{equation}
    Dice = \frac{2 \times\left | X\bigcap Y \right |}{\left |X  \right | + \left |Y  \right |}
\end{equation}

Here X is the set of predicted pixels of a specific class from a model and Y is the pixels belonging to the ground truth. A higher dice score indicates higher segmentation quality.

\subsection{Evaluation Results}
\begin{table}[ht]
\centering
\begin{adjustbox}{width=0.98\linewidth}
\begin{tabular}{@{}cccccc@{}}
\toprule
\multirow{2}{*}{Model} & \multirow{2}{*}{Backbone} & \multicolumn{2}{c}{CUB(\%)} & \multicolumn{2}{c}{Waterbird(\%)} \\
            &           & BG    & Bird  & BG    & Bird  \\ \midrule
Mask2Former & Swin-T & 99.42 & 96.05 & \textbf{98.74} & \textbf{91.84} \\
Mask2Former & ResNet50  & 99.43 & \textbf{96.12} & 98.72 & 91.81 \\ \bottomrule
\end{tabular}
\end{adjustbox}
\caption{Evaluation Results - Binary Segmentation} 
\label{tab:results_binary_seg}
\end{table}

The results of the evaluation are given in \autoref{tab:results_binary_seg}. From the table, we see that the model generalizes very well to both the in-distribution CUB and OOD Waterbirds test set. 

We chose the model with the Swin-Tiny backbone as it performed better overall.

\end{document}